\documentclass[a4paper,fleqn]{cas-dc}

\usepackage{algorithm}
\usepackage{algpseudocode}
\usepackage[numbers]{natbib}
\usepackage{multirow}
\usepackage{makecell}
\usepackage{xurl}
\def\tsc#1{\csdef{#1}{\textsc{\lowercase{#1}}\xspace}}
\tsc{WGM}
\tsc{QE}
\begin{document}
\let\WriteBookmarks\relax
\def\floatpagepagefraction{1}
\def\textpagefraction{.001}

% Short title - abbreviated version for running headers
% Short title - abbreviated version for running headers
\shorttitle{Diagnosing Domain Shift in Android Malware Detection}

% Short author - abbreviated author list for running headers
\shortauthors{Islam and Khan}

% Main title of the paper
\title [mode = title]{Diagnosing and Mitigating Domain Shift in Permission-Based Android Malware Detection}

% Title footnote mark
\tnotemark[1]

% Title footnote 1.
\tnotetext[1]{This research did not receive any specific grant from funding agencies in the public, commercial, or not-for-profit sectors.}

% --- FIRST AUTHOR (Corresponding Author) ---
\author[1]{Md Rafid Islam}[orcid=0009-0004-1140-7962]
\cormark[1]
\ead{md.islam.241@northsouth.edu}
\credit{Conceptualization, Investigation, Methodology, Software, Validation, Formal analysis, Writing - Original draft preparation, Writing - Review \& Editing, Visualization}

% --- SECOND AUTHOR ---
\author[1]{Mohammad Ashrafuzzaman Khan}
\ead{mohammad.khan02@northsouth.edu}
\credit{Supervision, Writing - Review \& Editing}

% --- AFFILIATION ---
\affiliation[1]{
    organization={Department of Electrical and Computer Engineering, North South University},
    addressline={Bashundhara},
    city={Dhaka},
    postcode={1229},
    country={Bangladesh}
}

% Corresponding author text
\cortext[1]{Corresponding author}

\begin{abstract}
Machine learning-based Android malware detectors often fail in real-world deployment due to domain shift, where models trained on one data source perform poorly on applications from another. This paper presents a study on the generalizability and interpretability of permission-based detectors under cross-domain conditions. Using two complementary datasets (PerMalDroid and NATICUSdroid) and five ensemble classifiers, we establish intra-domain baselines of 89--90\% accuracy on PerMalDroid under independent outer-fold evaluation and approximately 97\% on NATICUSdroid, and then quantify asymmetric performance degradation under cross-domain evaluation. Models trained on PerMalDroid show a moderate ability to generalize to NATICUSdroid (85--87\% mean accuracy), whereas the performance in the opposite direction is only 69--73\%. A matched ablation study shows that feature reduction improves mean transfer accuracy for four of five classifiers. Complementary PCA, nearest-neighbor, and SHAP analyses provide evidence of cross-domain class overlap and differences in permission-based predictive patterns. Supervised hybrid training using common features achieves 85--88\% accuracy on PerMalDroid and approximately 97\% on NATICUSdroid, improving over source-only transfer but not outperforming target-only training. CORAL-based adaptation aligns covariance structures, but its outcomes depend on the classifier and regularization setting. Modest gains are observed for boosting classifiers in the harder direction, indicating potential benefits when unlabeled target samples are available. These findings highlight the importance of independent feature selection, interpretability, and cross-domain evaluation for improving permission-based detectors.
\end{abstract}

% Keywords
% Each keyword is seperated by \sep

\ExplSyntaxOn
\RenewDocumentCommand \dashrule { O{.4pt} m m }
  {
    \color{black!50}
    \skip_vertical:n { #2 }
    \noindent \rule { \columnwidth } { #1 }
    \normalcolor \skip_vertical:n { #3 }
  }
\cs_set:Npn \__first_head:
  { \parbox[t]{\columnwidth}{\color{black!20}\rule{\columnwidth}{0pt}} }
\cs_set:Npn \__first_foot:
  { \parbox[t]{\columnwidth}{\rule{\columnwidth}{.2pt}\\ \__first_footerline: \hfill Page~ \thepage{} ~of~ \lastpage} }
\ExplSyntaxOff

\begingroup
\makeatletter
\let\Hy@Warning\@gobble
\makeatother
\maketitle
\endgroup

% Main text
\section{Introduction}

Android malware has become more sophisticated due to AI-driven adaptability, advanced evasion, and multi-vector exploitation [1]. Permissions play a critical role in this evolution by providing malware with the means to escalate privileges, operate stealthily, and execute complex attacks [2]. In essence, permissions are rules that restrict access to specific device features or data. While necessary for legitimate app functionality, malicious apps frequently exploit them. Prominent examples from 2023 include the ``Xamalicious'' malware, which abused Android accessibility privileges to enable device takeover and potential ad fraud [3], and the ``SpinOk'' spyware, which misused external storage permissions granted to its host applications to access user files [4]. Therefore, analyzing permissions is a highly effective method for detecting Android malware.

\begin{figure}[pos=t]
\centering
\includegraphics[width=\columnwidth]{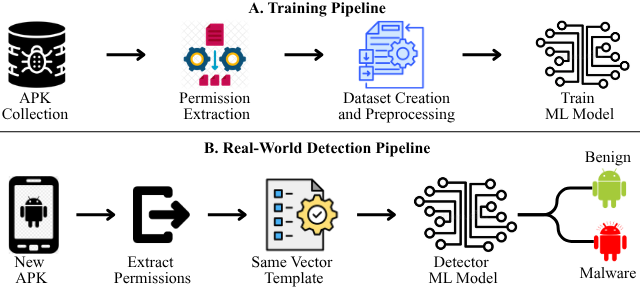}
\caption{Training machine learning models to detect Android malware using permissions as features.}
\label{fig:pipeline}
\end{figure}

Figure~\ref{fig:pipeline} illustrates a typical permission-based malware detection pipeline. First, benign and malicious apps are collected and analyzed to extract manifest permissions, which are mapped onto binary feature vectors. This standardized vector representation enables machine learning models to learn patterns that differentiate malicious from benign applications. When a new APK arrives, the deployed detector turns the requested permissions into the same binary vector used during training and outputs a malware or benign judgment based solely on its permission profile.

While machine learning (ML)-based detectors have demonstrated high intra-domain accuracy, their robustness across heterogeneous environments remains uncertain and underexplored. In practice, models trained on one domain- an exact environment, characteristics, and statistical distribution of the Android application ecosystem from which the data is drawn often fail when applied to another, a phenomenon known as the domain shift problem, which is identified as a critical open challenge in comprehensive surveys of Android malware detection [5, 6]. This issue arises because permission-based Android malware datasets vary widely in feature coverage and app distribution. Moreover, even when the same permissions appear in both domains, their relationship to the class label may differ fundamentally, a phenomenon known as concept shift.

Even though permission-based malware detection is quite developed, the cross-domain generalizability of these models has not been systematically investigated. A detector that works well on the distribution it has been trained on but fails when applied to apps from a different source, market, or collection period gives only limited confidence for real-world use. Moreover, even in cases where high accuracy has been reported, the reasons for the model's decisions have seldom been looked at in a cross-domain context; to understand why a model fails across domains, it is necessary to use interpretability tools that go beyond just looking at overall performance metrics. In order to fill these gaps, we carry out a systematic cross-domain study making use of two datasets with different structures: the curated permission-based Android malware dataset (PerMalDroid) used in [7] and the NATICUSdroid dataset [8]. Together, these datasets offer a controlled yet realistic environment for diagnosing and counteracting domain shift in permission-based detection. The study is based on the following main objectives:

\begin{itemize}
\item To establish a rigorous intra- and cross-domain performance benchmark for permission-based ensemble classifiers, quantifying the extent and asymmetry of domain shift between independently curated datasets.

\item To investigate factors associated with cross-domain performance degradation through feature set analysis, high-dimensional nearest-neighbor analysis, and an explainable AI framework.

\item To evaluate practical mitigation strategies and assess their effectiveness and limitations in recovering cross-domain generalization.
\end{itemize}

Through extensive experiments with five state-of-the-art tree-based ML models (Random Forest, LightGBM, XGBoost, Histogram-Based Gradient Boosting, and CatBoost), we demonstrate the effect of domain shift on classifier performance and conduct a comprehensive study of this problem. Our contributions are summarized as follows:

\begin{itemize}
\item An empirical benchmark that measures the asymmetric domain shift among five ensemble classifiers, with the Pd feature selection performed separately from the final intra-domain test folds. An ablation study is carried out to examine the impact of feature reduction on cross-domain transfer, while repeated source-model fits assess sensitivity to source-data partitioning.

\item A diagnostic analysis combining PCA visualization, high-dimensional nearest-neighbor analysis, and SHAP interpretability, providing evidence of cross-domain class overlap, feature set mismatch, and changes in permission-based predictive patterns associated with transfer failure.

\item An evaluation of supervised hybrid training on common permissions, achieving 85--88\% accuracy on PerMalDroid and approximately 97\% on NATICUSdroid. Matched source-only, target-only, and size-matched controls assess the contribution of labeled target data and training size. An evaluation of unsupervised CORAL adaptation examines classifier- and regularization-dependent outcomes, with paired bootstrap intervals quantifying uncertainty in the control comparisons.
\end{itemize}

\section{Related Works}

Several studies have focused on the static analysis of permissions used to develop malware classifiers. Peiravian and Zhu [9], for example, combined permissions with API calls to train ML models and produced promising results in identifying malware. Rovelli and Vigfússon [10] suggested a lightweight detection technique predicated exclusively on app permissions specified in the manifest file, achieving a detection rate of 92--94\% for previously unobserved malware. Milosevic \emph{et al.} [11] introduced two ML-assisted methodologies for the static analysis of Android malware, one of which relies on permissions and is computationally inexpensive. The evaluation of the permission-based classification models shows an F-measure of 89\%. The significance of a scalable malware detection strategy has been underlined by Li \emph{et al.} [12], who presented a detection system based on permission usage analysis that classifies various malware and benign app families using ML-based techniques. They identified only 22 significant permissions and conducted a comparison between these 22 permissions and all others. Their proposed approach was able to achieve 91.4\% accuracy in detecting new malware samples. Similarly, using a permission-based approach, Ilham \emph{et al.} [13] classified applications as either benign or malicious and highlighted that feature selection is crucial, as using all permissions leads to noise and overfitting.

More recent efforts, such as those by Kim \emph{et al.} [14], introduced a methodology that integrates both custom and built-in permissions, established four categories of permission sets, and evaluated seven classifiers on the twenty largest malware families from the DREBIN dataset. LightGBM and AdaBoost demonstrated satisfactory results. Furthermore, Mathur \emph{et al.} [8] focused on identifying malicious applications by analyzing both native and custom permissions requested by apps, and their experimental results showed that the Random Forest achieved the best result, with 97\% accuracy. Herron \emph{et al.} [15] examined the efficacy of Random Forest, SVM, Naive Bayes, and K-Means using manifest permissions for categorization, and Random Forest achieved the highest accuracy at 81.5\%. In addition, Akbar \emph{et al.} [16] demonstrated that reducing the feature set to just 5 items: 3 key permissions and 2 calculated metrics (permission rate and code size) could maintain accuracy near 90\% while improving computational efficiency. Arif \emph{et al.} [17] furthered malware detection research by experimenting with a dataset of 5,000 Drebin malware and 5,000 benign samples from AndroZoo, using feature selection to pick the most discriminative permissions, and testing several classifiers. Random Forest outperformed the other algorithms with a 91.6\% true positive rate (TPR) and the best overall accuracy.

Şahin \emph{et al.} [18] employed a linear regression-based feature selection strategy to lower the dimension of the feature vector and achieved an F-measure of 0.961 using at least 27 features. Mahindru \emph{et al.} [19] developed the PermDroid framework that employed feature selection techniques like t-tests and univariate regression to prioritize relevant permissions to improve model efficiency. Finally, Mawoh \emph{et al.} [20] introduced \textit{SigColDroid}, an innovative permission-based ML approach to differentiate between benign Android apps, single-app malware, and colluding malware, using a balanced dataset of 1,455 apps. Among the evaluated classifiers, the highest ROC-AUC (99.48\%) was achieved by Random Forest, and superior overall metrics (96.91\% accuracy and 96.90\% F1-score) by LightGBM. These works are summarized in Table~\ref{tab:literature_review} to provide a consolidated overview.

\begin{table*}[t]
\centering
\caption{Summary of Permission-Based Android Malware Detection Literature}
\label{tab:literature_review}
\begin{tabular}{p{2cm} p{1.5cm} p{1.9cm} p{1.6cm} p{4.3cm} p{3.5cm}}
\hline
\textbf{Reference} & \textbf{Framework} & \textbf{Feature(s)} & \textbf{Dataset Size (Apps)} & \textbf{Feature Selection / Novelty} & \textbf{Key Result} \\
\hline
Peiravian \& Zhu (2013) [9] & -- & Perms + API Calls & 1,860 & Early combination of static features for ML & 96.39\% Accuracy, 94.9\% Precision, 94.1\% Recall \\
\hline
Rovelli \& Vigfússon (2014) [10] & PMDS & Perms & 2,950 & Permissions as behavioral markers; zero-day detection focus & 92--94\% Detection Rate, 1.52--3.93\% FPR \\
\hline
Milosevic et al. (2017) [11] & Seraphim-droid & Perms & 400 & Computationally inexpensive; deployed in a live app & 89\% F-measure \\
\hline
Li et al. (2018) [12] & SigPID & Perms & 10,988 & Three-level pruning to identify only 22 significant permissions & SVM: ~90\% Accuracy, (4--32x faster) \\
\hline
Ilham et al. (2018) [13] & -- & Perms & 731 & Filter feature selection algorithms & SMO: 98.08\% Accuracy \\
\hline
Herron et al. (2021) [15] & -- & Perms & 5,243 & Evaluation of classic ML models on manifest perms & Random Forest: 81.5\% Accuracy, 82.5\% Precision \\
\hline
Arif et al. (2021) [17] & -- & Perms & 10,000 & Feature selection on Drebin \& Androzoo datasets & Random Forest: 91.6\% TPR \\
\hline
Mathur et al. (2021) [8] & NATICUS-droid & Native + Custom Perms & 29,332 & Uses trend analysis (2010--2019) to select relevant perms & Random Forest: 97\% Accuracy, 0.96 F1-score \\
\hline
Kim et al. (2021) [14] & -- & Built-in + Custom Perms & 5,560 & Malware Family Classification; focus on imbalanced metrics & LightGBM: ~96.9\% Accuracy \\
\hline
Akbar et al. (2022) [16] & PerDRaML & Perms + Smali size + Perm Rate & 10,000 & Multi-level methodology; aims for symmetry in permissions & Random Forest: 89.96\% Accuracy \\
\hline
Şahin et al. (2023) [18] & -- & Perms & 2,000 & Linear Regression-based feature selection (27 features) & F-measure: 96.1\% \\
\hline
Mahindru et al. (2024) [19] & PermDroid & Perms & 500,000 & Two-stage feature selection (t-test, LR, multivariate) on 500K apps & Deep Neural Network: 98.8\% Accuracy \\
\hline
Mawoh et al. (2025) [20] & SigColDroid & Perms + Smali size + Perm Rate & 1,455 & Detects colluding apps vs. single-app malware; identifies top 5 permissions & Random Forest: 0.99 AUC, LightGBM: 96.91\% Accuracy \\
\hline
\end{tabular}
\end{table*}

Although the aforementioned contributions have advanced the use of permissions in malware detection, a common drawback is the lack of rigorous cross-domain evaluation on data with different distributions. A number of studies base their evaluations on data from a similar source or a similar time period, or they do not test the models' ability to generalize across both the benign and malicious domains at the same time, thereby limiting the models' usefulness in a wide range of real-world situations. In particular, no prior work in permission-based Android malware detection has evaluated model performance across two independently curated datasets from distinct sources. Cross-domain generalization remains underexplored in permission-based detection, while comprehensive surveys highlight the challenges of maintaining malware detection performance when faced with different data sources or environments [5, 6].Moreover, interpretability in these models, particularly using tools like SHAP to explain feature contributions, has not been systematically applied to cross-domain settings. According to Zhang \emph{et al.} [21], despite the effectiveness of AI-based malware detection techniques, they operate in a black-box fashion, making it difficult to understand the logic behind a conclusion. Our work addresses these gaps through a systematic cross-domain evaluation and a comprehensive interpretability analysis.

\section{Methodology}

\subsection{Dataset Description}

PerMalDroid (Pd) was compiled from multiple open-source platforms and research repositories [7]. It consists of 1,168 samples (566 benign and 602 malware) and 942 permissions. The second dataset, NATICUSdroid (Nd) Android Permissions Dataset, introduced by Mathur \emph{et al.} [8], is a large-scale dataset containing 86 permissions and 29,332 samples, with an almost even split between 14,700 malware from Argus Lab's AMD and 14,632 benign applications from Androzoo.

\subsection{Feature Selection using Pearson Correlation}

No feature selection was applied to Nd because its feature space already contains only 86 permissions. For Pd, a univariate filter-based approach using the Pearson correlation coefficient was employed. The coefficient $r$ between a feature vector $X$ and the target vector $Y$ is defined as:
\begin{equation}
r = \frac{\sum_{i=1}^{n}(X_i - \bar{X})(Y_i - \bar{Y})}
{\sqrt{\sum_{i=1}^{n}(X_i - \bar{X})^2}
\sqrt{\sum_{i=1}^{n}(Y_i - \bar{Y})^2}}
\label{eq:pearson}
\end{equation}
where $X_i$ and $Y_i$ are the $i$-th observations, $\bar{X}$ and $\bar{Y}$ are their means, and $n$ is the number of samples. Permissions were ranked by their absolute correlation with the class label, and the top-$k$ permissions were selected.

Constant permissions were removed, and the remaining permissions were ranked using only the inner training subset in each outer iteration. We used five-fold stratified outer cross-validation with an internal hold-out validation split in order to separate feature selection from final performance evaluation. In each outer iteration, four folds formed the outer training partition, while the remaining fold was reserved exclusively for final intra-domain testing. The outer training partition was further divided into stratified inner training and validation subsets in an 80:20 ratio.

For each classifier, a full-feature reference model was trained using all permissions that were non-constant in the inner training subset, and its accuracy was measured on the inner validation subset. Candidate models were then trained using increasing numbers of permissions ranked by absolute Pearson correlation computed on the inner training subset. The smallest feature count whose validation accuracy matched or exceeded that of the full-feature reference model was selected.

After selection, the chosen permission set was fixed. A fresh classifier was trained on the entire outer training partition. It was then evaluated on the untouched outer Pd test fold and, separately, on Nd for cross-domain evaluation. Neither the outer test labels nor Nd labels were used for feature selection. The procedure was repeated across all five outer folds and evaluation metrics were reported as mean $\pm$ standard deviation. The selected feature count was allowed to vary across folds and classifiers, reflecting the selection criterion's dependence on the source training and validation samples. 

\subsection{Ensemble Classifiers}

This study evaluates five classifiers: Random Forest (RF), XGBoost (XGB) [22], LightGBM (LGBM) [23], CatBoost (CB) [24], and Histogram-based Gradient Boosting (HGB). The selection of these models is based on their established performance in prior work and their theoretical suitability for the task. Empirical evidence from related work highlights tree-based ensembles, in particular RF [8, 15, 16, 17] and LGBM [14, 20], as top performers for permission-based detection. On this basis, we included modern gradient-boosting variants (XGB, CB, HGB) for a comprehensive comparison. We evaluated these models using standard binary classification metrics: accuracy, precision, recall, F1-score, and AUC.

\subsection{Experimental Setup and Evaluation}

\begin{figure}[pos=t]
\centering
\includegraphics[width=\columnwidth]{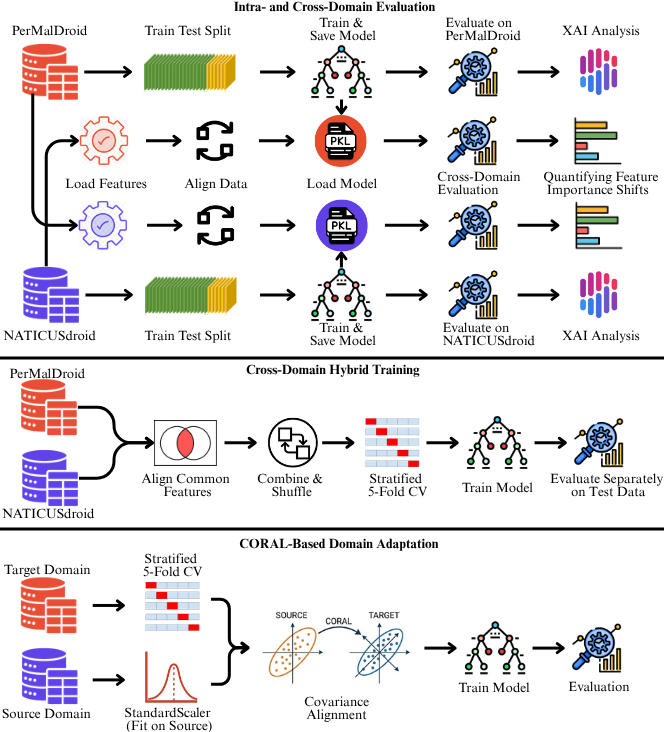}
\caption{Simplified overview of the experimental workflow showing intra- and cross-domain evaluation, explainable AI analysis, hybrid training evaluation, and CORAL-based domain adaptation.}
\label{fig:workflow}
\end{figure}

The experiments were organized into several categories to comprehensively assess model performance: intra-domain evaluation, cross-domain evaluation, hybrid training, and CORAL-based domain adaptation. The workflow is simplified and summarized in Figure~\ref{fig:workflow}.

\subsubsection{Intra- and Cross-Domain Evaluation}

First, an intra-domain evaluation was conducted to establish a baseline performance level for each model on its native dataset. For Pd, five-fold stratified outer cross-validation was used, with feature selection performed within each outer training partition as described above. For Nd, models were trained and evaluated using a single stratified train-test split (80--20\%). The trained models were saved for subsequent cross-domain testing, in which Pd-trained models were tested directly on Nd and vice versa, denoted as Pd $\rightarrow$ Nd and Nd $\rightarrow$ Pd. Pd intra-domain and Pd $\rightarrow$ Nd results were reported as mean $\pm$ standard deviation across the five outer iterations. Missing features were added as zero-filled columns to make sure that the training and test feature spaces were compatible. The order of the features was exactly matched to that of the training data, and any extra features that were present only in the test dataset were disregarded. The models were then applied to the aligned datasets for the purpose of evaluation.

\subsubsection{Cross-Domain Hybrid Training}

Supervised hybrid training was evaluated using the 85 permissions common to Pd and Nd. For each transfer direction, five-fold stratified cross-validation was applied to the target dataset, with each target-test fold excluded from training. Four conditions were compared using identical classifier settings and target-test folds: source-only training, target-only training, full hybrid training using all source samples and the target-training partition, and size-matched hybrid training.

The size-matched hybrid contained the same total number of samples as target-only training. The source contribution was the smaller of half this total (rounded down) and the available source sample count; the remaining samples were drawn from the target-training partition. Samples were selected without replacement using stratification by class, and combined training data were shuffled. Metrics were reported as mean $\pm$ standard deviation across target folds. Unlike direct transfer and CORAL, hybrid training used target-domain labels to train the classifier.

\subsubsection{CORAL-Based Domain Adaptation}

Correlation Alignment (CORAL) was employed as an unsupervised domain adaptation technique that aligns the second-order statistics of source and target feature spaces [25]. Both datasets were restricted to their 85 common permissions, and all source-domain samples were used for training.

The target dataset was divided into five parts using five-fold stratified cross-validation. In each of these parts, 80\% of the target samples were used to estimate the alignment, while the other 20\% were set aside for evaluation. The target labels were used for the purpose of stratification and for evaluation, but not for alignment or for training the classifier.

Feature standardization parameters were estimated using only source-domain samples and applied to both domains. Let $X_s$ and $X_t$ denote the standardized source and target-training feature matrices, respectively. Their regularized covariance matrices were computed as:
\begin{equation}
C_s = \operatorname{cov}(X_s) + \lambda I
\label{eq:cov_source}
\end{equation}
\begin{equation}
C_t = \operatorname{cov}(X_t) + \lambda I
\label{eq:cov_target}
\end{equation}
where $I$ denotes the identity matrix and $\lambda$ controls covariance regularization. The original setting was $\lambda = 10^{-3}$, and sensitivity was evaluated at $\lambda \in \{10^{-4},10^{-3},10^{-2},10^{-1},1\}$ without selecting a value based on target-test performance.

CORAL transforms the source features according to:
\begin{equation}
X_s' = (X_s - \mu_s) C_s^{-1/2} C_t^{1/2} + \mu_t
\label{eq:coral_transform}
\end{equation}
where $\mu_s$ and $\mu_t$ are the source and target-training means. The matrix powers were computed using symmetric eigendecomposition.

Each classifier was trained on $(X_s',y_s)$ and evaluated on the unseen target-test partition. A matched no-CORAL baseline was trained on the standardized source data using identical source samples, common features, classifier settings, and target-test folds. Metrics were reported as mean $\pm$ standard deviation across target folds. Algorithm~\ref{alg:coral} summarizes the procedure.

\begin{algorithm}[t]
\caption{CORAL-Based Domain Adaptation and Sensitivity Analysis}
\label{alg:coral}
\begin{algorithmic}[1]
\Require Source dataset $(X_s,y_s)$, target dataset $(X_t,y_t)$
\Ensure Mean $\pm$ standard deviation of evaluation metrics
\State Restrict both datasets to common permissions
\State Standardize features using source-domain statistics
\State Create 5 stratified target folds using $y_t$
\For{each classifier}
    \State Train no-CORAL baseline using $(X_s,y_s)$
\EndFor
\For{each target fold}
    \State Split target data into $X_t^{train}$ and $X_t^{test}$
    \State Evaluate baselines on $X_t^{test}$ and record metrics
    \For{$\lambda \in \{10^{-4},10^{-3},10^{-2},10^{-1},1\}$}
        \State Compute $C_s$ and $C_t$ using Eqs.~\ref{eq:cov_source}--\ref{eq:cov_target}
        \State Compute $X_s'$ using Eq.~\ref{eq:coral_transform}
        \For{each classifier}
            \State Train a fresh classifier using $(X_s',y_s)$
            \State Evaluate on $X_t^{test}$ and record metrics
        \EndFor
    \EndFor
\EndFor
\State Aggregate metrics across folds for each classifier and setting
\State \Return evaluation metrics (mean $\pm$ standard deviation)
\end{algorithmic}
\end{algorithm}

The 95\% confidence intervals for pooled accuracy and malware recall with respect to the CORAL and hybrid controls were obtained by means of 10,000 bootstrap resamples, the resampling being stratified by target fold and class. When evaluating the accuracy differences, paired resampling of the predictions on the same target samples was carried out. The pointwise intervals obtained in this way have not been adjusted for multiple comparisons and are based on the fitted models and the folds used; they do not take into account the variability due to retraining or the uncertainty arising from different data sources. While the bootstrap estimates average the target predictions across the folds, the main performance tables give the unweighted fold means together with the standard deviations. Detailed numerical confidence intervals are given in Appendix~\ref{app:confidence_intervals}.

The control baselines use all source samples and the 85 common permissions, whereas direct-transfer evaluation uses the source training partitions and feature spaces described above. Their results therefore represent different experimental settings. 

\subsection{SHAP for Model Interpretability}

We employed SHapley Additive exPlanations (SHAP), a unified, game-theoretic approach that quantifies the contribution of each feature (permission) [26]. SHAP enables the identification of critical permissions that drive malware classification and offers explanations for specific predictions. For a model $f$ and instance $x$, the SHAP value $\phi_i$ for feature $i$ is calculated as:
\begin{equation}
\phi_i(f,x) = \sum_{S \subseteq F \setminus \{i\}} \frac{|S|! (|F| - |S| - 1)!}{|F|!} [f(S \cup \{i\}) - f(S)]
\label{eq:shap}
\end{equation}
where $F$ is the set of all features, $S$ is a subset of features excluding $i$, and $f(S)$ is the prediction from the model trained only on the feature subset $S$. The summation is over all possible subsets $S$.

SHAP provides both global and local interpretability by attributing each prediction to individual features in a manner consistent with cooperative game theory. RF was used as the representative model for global interpretability due to its strong performance and widespread use in related works. By combining global insights (feature importance distributions) with local reasoning (instance-level justifications), the SHAP analysis provides a complete interpretability framework.

\section{Experimental Results and Analysis}

The results and analysis are presented in a narrative flow, beginning with the baseline intra-domain performance, followed by cross-domain evaluation, scrutiny of domain shift, XAI insights, and concluding with the outcomes of the mitigation strategies.

\subsection{Intra-Domain Performance}

Intra-domain baseline results for the Pd and Nd datasets are reported in Tables~\ref{tab:permaldroid} and~\ref{tab:naticusdroid}, respectively. For Pd, selected-feature results are reported in Table~\ref{tab:permaldroid}, while matched full-feature comparisons are presented in Table~\ref{tab:ablation_study}.

\subsubsection{PerMalDroid Dataset}

With feature selection performed independently of the outer test folds, models achieved mean accuracies of 89.12--90.41\% and AUC values of 0.95--0.96 (Table~\ref{tab:permaldroid}). XGB achieved the highest mean accuracy (90.41\%), while RF showed the lowest accuracy standard deviation (2.03 percentage points). Selected feature counts varied across folds and classifiers, ranging from 23 to 559, indicating that the feature subset required to match full-feature validation accuracy is split-dependent. In the remainder of this paper, Pd intra-domain performance refers to these independent outer-fold results.

\begin{table}[pos=t]
\caption{Performance Summary on PerMalDroid. Metrics are reported as mean $\pm$ standard deviation across five outer folds; precision, recall, and F1-score refer to malware.}
\label{tab:permaldroid}
\centering
\small
\setlength{\tabcolsep}{3pt}
\renewcommand{\arraystretch}{1.4}
\begin{tabular*}{\tblwidth}{@{} LLLLLL @{}}
\toprule
\textbf{Metric} & \textbf{RF} & \textbf{XGB} & \textbf{LGBM} & \textbf{CB} & \textbf{HGB} \\
\midrule
Feature range & 28--290 & 32--252 & 36--559 & 29--58 & 23--152 \\
Accuracy (\%) &
\makecell{$89.90$\\$\pm 2.03$} &
\makecell{$90.41$\\$\pm 2.96$} &
\makecell{$89.81$\\$\pm 3.62$} &
\makecell{$89.29$\\$\pm 2.46$} &
\makecell{$89.12$\\$\pm 3.79$} \\
Precision (\%) &
\makecell{$90.21$\\$\pm 3.53$} &
\makecell{$91.94$\\$\pm 1.48$} &
\makecell{$90.27$\\$\pm 2.24$} &
\makecell{$91.62$\\$\pm 1.19$} &
\makecell{$91.29$\\$\pm 3.16$} \\
Recall (\%) &
\makecell{$90.37$\\$\pm 3.61$} &
\makecell{$89.20$\\$\pm 5.43$} &
\makecell{$89.87$\\$\pm 5.36$} &
\makecell{$87.20$\\$\pm 4.86$} &
\makecell{$87.21$\\$\pm 5.27$} \\
F1-score (\%) &
\makecell{$90.21$\\$\pm 2.00$} &
\makecell{$90.49$\\$\pm 3.12$} &
\makecell{$90.04$\\$\pm 3.70$} &
\makecell{$89.30$\\$\pm 2.71$} &
\makecell{$89.17$\\$\pm 3.91$} \\
AUC &
\makecell{$0.957$\\$\pm 0.011$} &
\makecell{$0.958$\\$\pm 0.018$} &
\makecell{$0.957$\\$\pm 0.019$} &
\makecell{$0.955$\\$\pm 0.015$} &
\makecell{$0.951$\\$\pm 0.031$} \\
\bottomrule
\end{tabular*}
\end{table}

\subsubsection{NATICUSdroid Dataset}

Models trained and tested on the Nd dataset produced even stronger intra-domain results. Every model exceeded 96\% accuracy and achieved an AUC score of 0.99, demonstrating near-perfect classification. The precision and recall for all models were well-balanced at approximately 97\%, resulting in identical F1-scores for both the benign and malware classes. This minimal performance variation across different classifiers indicates that the Nd feature space is highly separable and capable of providing strong, consistent discriminatory signals for all the evaluated models.

\begin{table}[pos=t]
\caption{Performance Summary on NATICUSdroid. B and M denote benign and malware, respectively.}
\label{tab:naticusdroid}
\centering
\small
\setlength{\tabcolsep}{4pt}
\renewcommand{\arraystretch}{1.2}
\begin{tabular*}{\tblwidth}{@{} LLL @{}}
\toprule
\textbf{Metric} & \textbf{Range} & \textbf{Mean} \\
\midrule
Accuracy (\%) & 96.88--97.26 & 97.00 \\
Precision (B/M, \%) & 96--97 / 97--98 & 97 / 97 \\
Recall (B/M, \%) & 97--98 / 96--97 & 97 / 97 \\
F1-score (B/M, \%) & 97 / 97 & 97 / 97 \\
AUC & 0.99 & 0.99 \\
\bottomrule
\end{tabular*}
\end{table}

\subsection{Cross-Domain Evaluation}

Table~\ref{tab:permaldroid_to_naticus} presents the results when classifiers trained on Pd (with feature selection applied) were tested on Nd, and Table~\ref{tab:naticusdroid_to_permaldroid} summarizes the reverse scenario. 

\subsubsection{PerMalDroid to NATICUSdroid}

All models achieved mean accuracies above 84\% on Nd, with RF attaining the highest accuracy of $87.40 \pm 1.41\%$ (Table~\ref{tab:permaldroid_to_naticus}). Malware recall ranged from 78.61\% to 83.77\%, indicating that some malicious applications remained undetected, while malware precision ranged from 89.38\% to 91.73\%. Mean AUC values of 0.91--0.94 indicate that the models retained discriminatory power under domain shift. These results show that Pd $\to$ Nd performance persists when feature selection is performed independently of the final Pd test folds.

\begin{table}[pos=t]
\caption{PerMalDroid $\to$ NATICUSdroid Results. Metrics are reported as mean $\pm$ standard deviation across five source-model fits; precision, recall, and F1-score refer to malware.}
\label{tab:permaldroid_to_naticus}
\centering
\small
\setlength{\tabcolsep}{3pt}
\renewcommand{\arraystretch}{1.2}
\begin{tabular*}{\tblwidth}{@{} LLLLLL @{}}
\toprule
\textbf{Metric} & \textbf{RF} & \textbf{XGB} & \textbf{LGBM} & \textbf{CB} & \textbf{HGB} \\
\midrule
Accuracy (\%) &
\makecell{$87.40$\\$\pm 1.41$} &
\makecell{$85.18$\\$\pm 0.55$} &
\makecell{$85.34$\\$\pm 1.74$} &
\makecell{$85.97$\\$\pm 0.79$} &
\makecell{$84.82$\\$\pm 0.53$} \\
Precision (\%) &
\makecell{$90.39$\\$\pm 0.46$} &
\makecell{$90.42$\\$\pm 1.69$} &
\makecell{$89.38$\\$\pm 0.94$} &
\makecell{$91.73$\\$\pm 1.41$} &
\makecell{$89.95$\\$\pm 2.28$} \\
Recall (\%) &
\makecell{$83.77$\\$\pm 2.94$} &
\makecell{$78.83$\\$\pm 1.86$} &
\makecell{$80.28$\\$\pm 3.21$} &
\makecell{$79.19$\\$\pm 2.42$} &
\makecell{$78.61$\\$\pm 2.96$} \\
F1-score (\%) &
\makecell{$86.93$\\$\pm 1.69$} &
\makecell{$84.20$\\$\pm 0.67$} &
\makecell{$84.57$\\$\pm 2.08$} &
\makecell{$84.97$\\$\pm 1.06$} &
\makecell{$83.83$\\$\pm 0.86$} \\
AUC &
\makecell{$0.931$\\$\pm 0.013$} &
\makecell{$0.930$\\$\pm 0.011$} &
\makecell{$0.931$\\$\pm 0.009$} &
\makecell{$0.936$\\$\pm 0.008$} &
\makecell{$0.913$\\$\pm 0.018$} \\
\bottomrule
\end{tabular*}
\end{table}

\subsubsection{NATICUSdroid to PerMalDroid}

In this direction, performance dropped significantly compared to both the intra-domain baseline and the Pd \texorpdfstring{$\rightarrow$}{->} Nd direction, with accuracies ranging from 68.84\% (HGB) to 73.03\% (RF). A critical finding is the pronounced asymmetry in detection performance, and that is the models maintained high recall for benign applications (86--87\%) but showed poor recall for malware (52--60\%), which means they failed to detect a substantial portion of malicious samples. Malware precision remains decent (80--83\%), suggesting that when models do flag malware, they are usually correct.

\begin{table}[pos=t]
\caption{NATICUSdroid $\to$ PerMalDroid Results. B and M denote benign and malware, respectively.}
\label{tab:naticusdroid_to_permaldroid}
\centering
\small
\setlength{\tabcolsep}{4pt}
\renewcommand{\arraystretch}{1.2}
\begin{tabular*}{\tblwidth}{@{} LLLLLL @{}}
\toprule
\textbf{Metric} & \textbf{RF} & \textbf{XGB} & \textbf{LGBM} & \textbf{CB} & \textbf{HGB} \\
\midrule
Accuracy (\%) & 73.03 & 69.78 & 69.86 & 70.80 & 68.84 \\
Precision (B/M, \%) & 67/83 & 64/82 & 64/81 & 65/82 & 63/80 \\
Recall (B/M, \%) & 87/60 & 87/53 & 86/54 & 87/55 & 86/52 \\
F1-score (B/M, \%) & 76/70 & 74/65 & 74/65 & 74/66 & 73/63 \\
AUC & 0.83 & 0.78 & 0.77 & 0.80 & 0.77 \\
\bottomrule
\end{tabular*}
\end{table}
\subsubsection{Domain Shift Impact}

\begin{figure}[pos=t]
\centering
\includegraphics[width=\columnwidth]{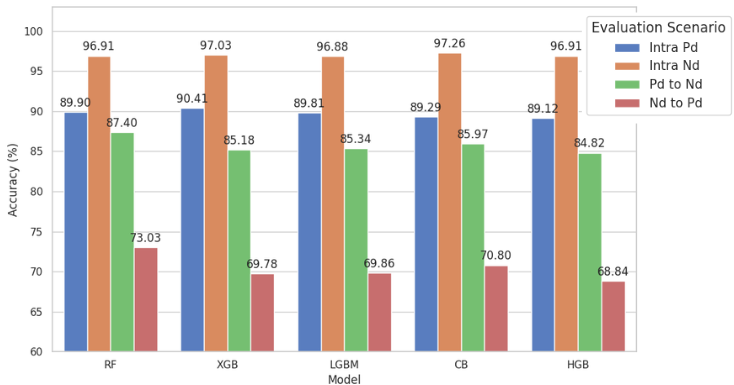}
\caption{Comparison of model accuracy across intra-domain and cross-domain evaluations.}
\label{fig:domain_shift}
\end{figure}

The analysis across domains showed asymmetric transfer performance (see Fig.~\ref{fig:domain_shift}). On the Pd dataset, the models obtained a mean intra-domain accuracy of about 89–90\%, whereas on Nd it was 97\%. However, when transferring from Pd to Nd, the mean accuracy remained at 85–87\%. By contrast, when transferring from Nd to Pd, the accuracy dropped to 69–73\%, which is about 24–28 percentage points less than the intra-domain accuracy baseline for Nd, and involved a particularly low rate of malware recall (as shown in Table~\ref{tab:naticusdroid_to_permaldroid}).

\begin{figure}[pos=t]
\centering
\includegraphics[width=\columnwidth]{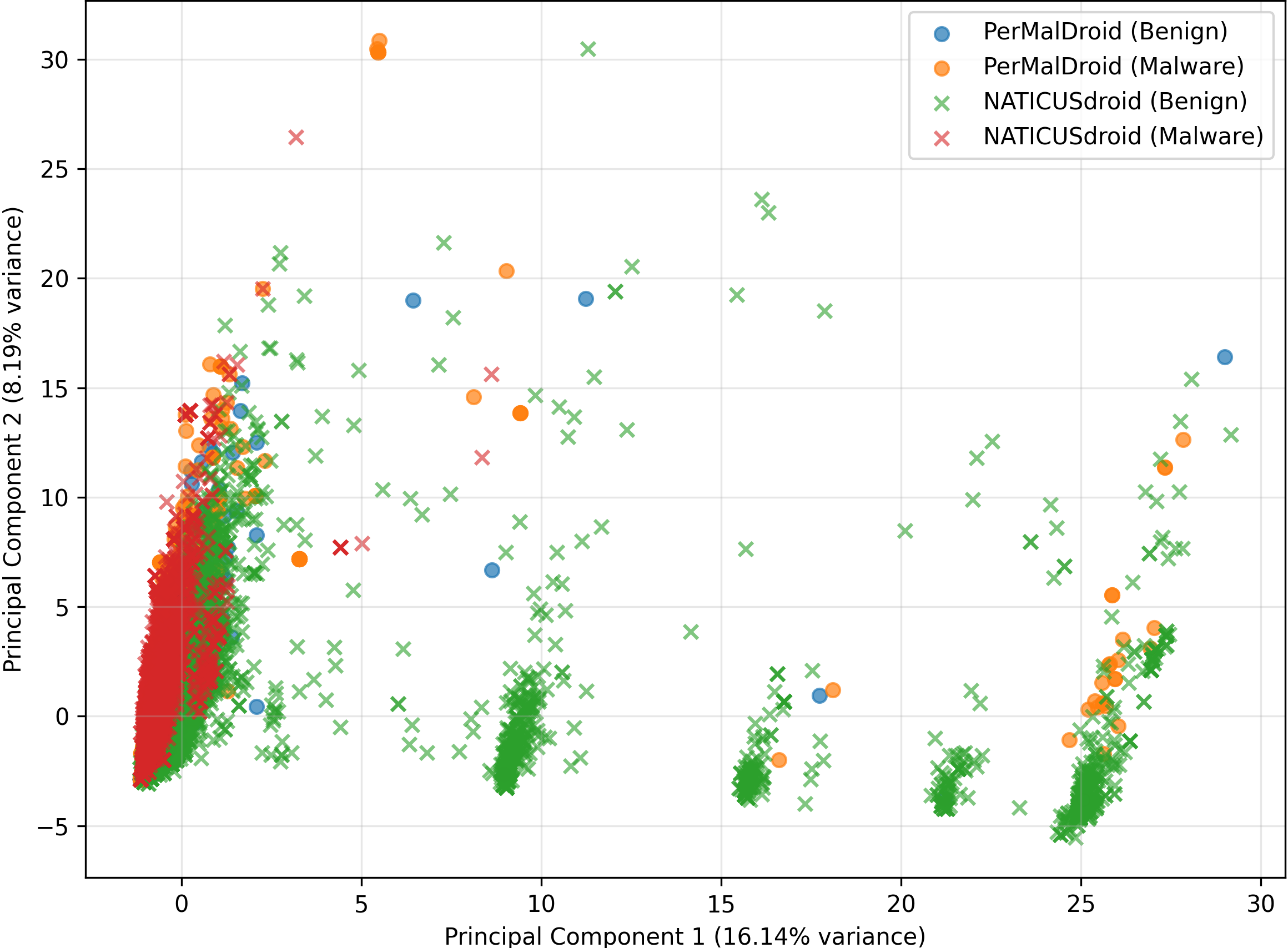}
\caption{PCA projection of permission-based feature vectors from Pd and Nd. PC1 and PC2 explain 16.14\% and 8.19\% of total variance, respectively (24.33\% combined).}
\label{fig:pca}
\end{figure}

The PCA projection (Fig.~\ref{fig:pca}) shows that there is an overlap between certain Pd malware and Nd benign samples. Yet the first two components account for only 24.59\% of the total variance, meaning that the apparent closeness in this projection does not prove there is overlap in the original feature space.

To investigate this pattern directly, an additional analysis used an Nd-trained RF model to identify misclassified Pd malware (false negatives) and correctly detected Pd malware (true positives). For each sample, nearest-neighbor Hamming distances to Nd benign and malware training samples were computed in the model's original 86-dimensional permission space, using the same cross-domain feature alignment. The distance ratio was defined as the nearest benign distance divided by the nearest malware distance; values below one indicate greater proximity to benign samples.

As shown in Table~\ref{tab:nearest_neighbor}, 59.00\% of false negatives were closer to Nd benign samples, compared with 5.87\% of true positives. Their median distance ratios were 0.83 and 1.50, respectively. Both groups had the same median benign distance (0.0581), but false negatives had a larger median malware distance (0.0698 versus 0.0349). Thus, relative proximity to benign rather than malware profiles was associated with missed detections beyond the two-dimensional projection. These findings support cross-domain class overlap as a contributing explanation, without establishing it as the sole cause.

\begin{table}[pos=t]
\caption{Nearest-Neighbor Analysis of Pd Malware in the Nd-Trained RF Feature Space. Undefined $0/0$ ratios were excluded from ratio medians; infinite ratios were retained.}
\label{tab:nearest_neighbor}
\centering
\small
\setlength{\tabcolsep}{4pt}
\renewcommand{\arraystretch}{1.1}
\begin{tabular*}{\tblwidth}{@{} LLL @{}}
\toprule
\textbf{Metric} & \textbf{False Negatives} & \textbf{True Positives} \\
\midrule
Samples & 261 & 341 \\
Closer to benign (\%) & 59.00 & 5.87 \\
Closer to malware (\%) & 6.90 & 78.59 \\
Equal distance (\%) & 34.10 & 15.54 \\
Median benign distance & 0.0581 & 0.0581 \\
Median malware distance & 0.0698 & 0.0349 \\
Median distance ratio & 0.83 & 1.50 \\
Exact matches in both classes & 16 & 1 \\
\bottomrule
\end{tabular*}
\end{table}

\subsection{Predictive Feature Set Mismatch and Ablation Study}

The full set of permissions in Pd includes 85 of the 86 permissions that are in Nd, but the Pd feature sets that were selected differed from one classifier to another and across outer folds. The selected subsets consisted of between 23 and 559 permissions, 18 to 82 of which were also found in Nd. The number of permissions that were selected in Pd but not in Nd varied from 5 to 477. These discrepancies show a mismatch in feature coverage, even though the overlap figures by themselves do not tell us which of the missing permissions cause the transfer to fail.

To assess the effect of feature selection, full-feature and selected-feature models were trained on identical outer training partitions and evaluated on the same Pd test folds and Nd samples. Full-feature models used all non-constant features within each outer training partition (634--658 features). Table~\ref{tab:ablation_study} reports the matched results.

Feature selection increased mean Pd $\rightarrow$ Nd accuracy by 4.85 percentage points for RF, 2.11 for XGB, 0.53 for LGBM, and 2.62 for HGB. CB showed a small mean decrease of 0.11 percentage points. On the other hand, full-feature models achieved higher mean intra-domain accuracy for all classifiers. Thus, feature reduction improved average transfer accuracy for four of five classifiers, despite lower intra-domain accuracy. This supports a model-dependent trade-off between source-domain fit and cross-domain performance.

\begin{table*}[t]
\caption{Matched Full-Feature and Selected-Feature Ablation. Accuracy is reported in percent as mean $\pm$ standard deviation across five outer iterations; $\Delta$ denotes selected minus full accuracy in percentage points.}
\label{tab:ablation_study}
\centering
\small
\setlength{\tabcolsep}{5pt}
\renewcommand{\arraystretch}{1.1}
\begin{tabular}{lccccc}
\toprule
\multirow{2}{*}{\textbf{Classifier}} &
\multicolumn{2}{c}{\textbf{Pd Intra-Domain}} &
\multicolumn{2}{c}{\textbf{Pd $\rightarrow$ Nd}} &
\multirow{2}{*}{\textbf{$\Delta$ Transfer}} \\
\cmidrule(lr){2-3} \cmidrule(lr){4-5}
& \textbf{Full} & \textbf{Selected} &
\textbf{Full} & \textbf{Selected} & \\
\midrule
RF & $91.35 \pm 1.13$ & $89.90 \pm 2.03$ &
$82.55 \pm 1.83$ & $87.40 \pm 1.41$ & $+4.85 \pm 1.50$ \\
XGB & $91.52 \pm 1.81$ & $90.41 \pm 2.96$ &
$83.08 \pm 2.80$ & $85.18 \pm 0.55$ & $+2.11 \pm 2.81$ \\
LGBM & $91.95 \pm 1.59$ & $89.81 \pm 3.62$ &
$84.81 \pm 1.31$ & $85.34 \pm 1.74$ & $+0.53 \pm 0.94$ \\
CB & $89.64 \pm 2.41$ & $89.29 \pm 2.46$ &
$86.09 \pm 3.45$ & $85.97 \pm 0.79$ & $-0.11 \pm 3.10$ \\
HGB & $92.21 \pm 1.78$ & $89.12 \pm 3.79$ &
$82.21 \pm 2.22$ & $84.82 \pm 0.53$ & $+2.62 \pm 1.92$ \\
\bottomrule
\end{tabular}
\end{table*}

\subsection{Global Interpretability with SHAP}

\subsubsection{Global Feature Impact Distributions}

\begin{figure*}[!t]
\centering
\includegraphics[width=0.8\textwidth]{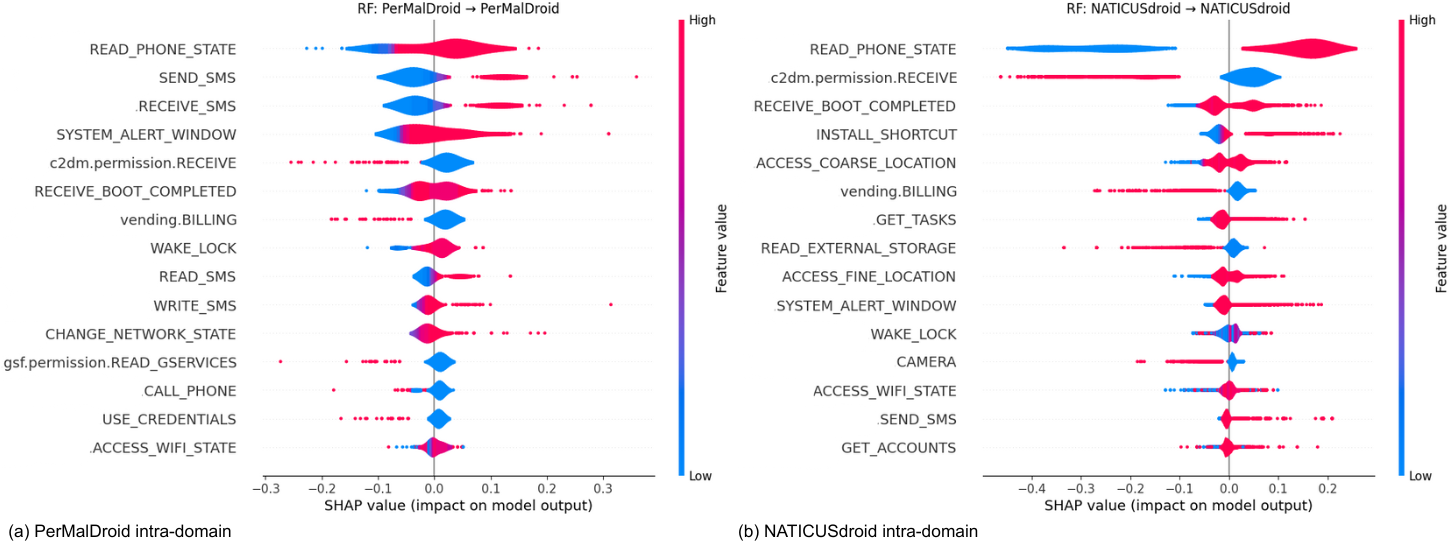}
\caption{SHAP violin plots showing the global feature impact distributions for the top 15 most important permissions in intra-domain evaluations.}
\label{fig:shap_global}
\end{figure*}

SHAP summary plots (violin plots) in \ref{fig:shap_global} visualize the impact of the top 15 most important features in the intra-domain evaluation scenarios. These plots reveal how each permission influences the model's output probability of a sample being malware. In these plots:

\begin{itemize}
\item The Y-axis lists Android permissions ranked by mean absolute SHAP value. The X-axis shows the SHAP value (positive = pushes toward malware, negative = toward benign).

\item Each point is a sample. The color indicates the feature's actual value for that sample: red = permission present (1), blue = absent (0).

\end{itemize}

Insights that can be drawn from these distributions:

\begin{itemize}

\item The horizontal distribution of points for each feature shows the range and density of its impacts. A wide distribution indicates the feature's effect varies significantly across different samples.

\item When points of different colors cluster on opposite sides, the feature provides a clean, consistent signal: the direction (which color leads to malware) depends on the permission's correlation with the target in that dataset.

\item When the same color appears on both sides of zero (e.g., red points on both positive and negative SHAP values), the feature's effect is context-dependent. The same permission can push toward malware in some apps but toward benign in others, depending on the combination of other permissions.

\item Bimodal distributions of this second type reveal that the model has learned nuanced, contextual rules rather than simple heuristics.
\end{itemize}

\subsubsection{Analysis of Key Permissions}

In both models, \texttt{READ\_PHONE\_STATE} had the highest mean absolute SHAP value. In Pd, the presence of this permission contributed to either class depending on the sample context. Whereas in Nd, its presence predominantly pushed predictions toward malware, and its absence toward benign. These contrasting attribution patterns indicate domain-dependent model reliance on the permission.

The model trained on Pd (Fig.~\ref{fig:shap_global}a) relies heavily on a set of permissions related to SMS, overlaid windows, and background execution. Permissions like \texttt{SEND\_SMS} and \texttt{RECEIVE\_SMS} exhibit strong bimodal distributions, which means that their presence can significantly increase the malware prediction score in some contexts, while having little effect or even a benign association in others. Similarly, \texttt{SYSTEM\_ALERT\_WINDOW} and \texttt{RECEIVE\_BOOT\_COMPLETED} show SHAP value distributions on both sides of zero, confirming that the classifier's interpretation of these permissions is highly context-dependent on other app characteristics.

In contrast, the model trained on Nd (Fig.~\ref{fig:shap_global}b) utilizes a different set of discriminative permissions. While \texttt{READ\_PHONE\_STATE}, \texttt{vending.BILLING}, and \texttt{RECEIVE\_BOOT\_COMPLETED} remain among the top features with distinctly bimodal distributions, a suite of data-access permissions emerged as uniquely important in this domain. These include \texttt{ACCESS\_COARSE} \texttt{\_LOCATION}, \texttt{ACCESS\_FINE\_LOCATION}, \texttt{CAMERA}, \texttt{GET\_ACCOUNTS}, and \texttt{READ\_EXTERNAL\_STORAGE}, which are not among the top predictors in the Pd-trained model. This difference in the ``fingerprint'' of important features between the two domains provides an explainable AI-driven reason for the cross-domain performance drop, and that is the models are relying on fundamentally different decision boundaries.

\subsection{Quantifying Cross-Domain Feature Importance Shifts}

\begin{figure*}[!t]
\centering
\includegraphics[width=0.8\textwidth]{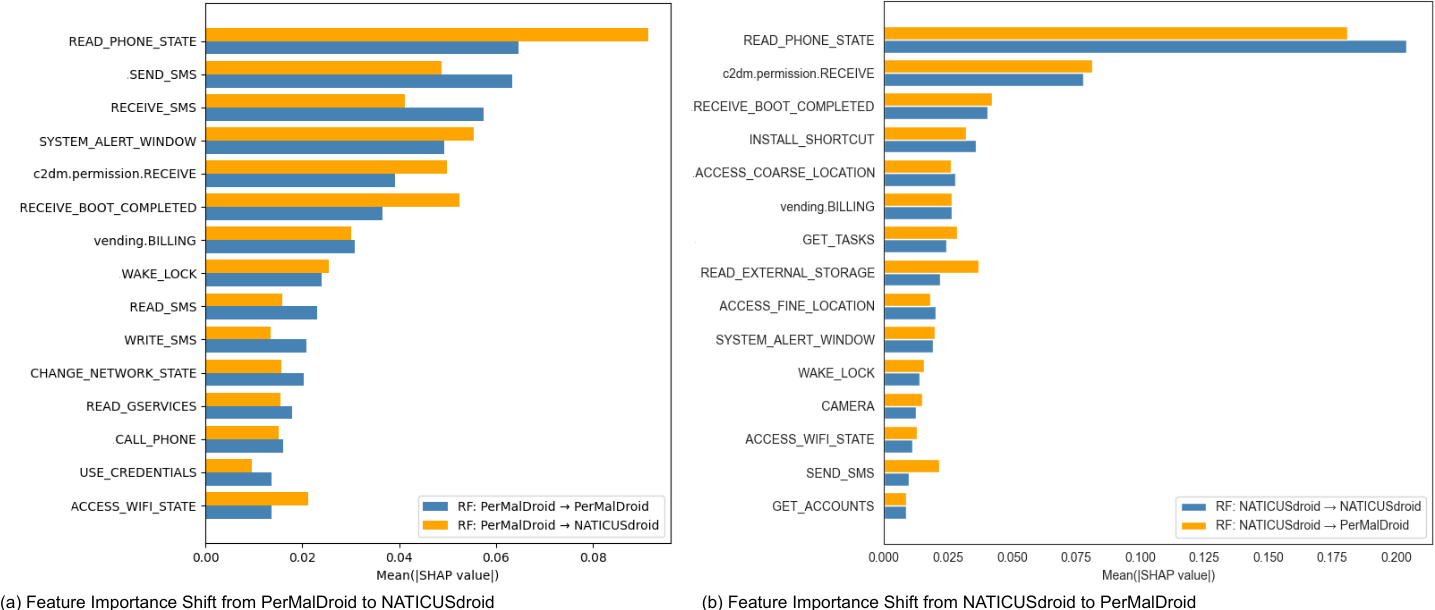}
\caption{Quantification of cross-domain feature importance shifts using mean absolute SHAP values. (a) Pd → Nd, (b) Nd → Pd.}
\label{fig:importance_shifts}
\end{figure*}

Building on the global interpretability analysis, we quantitatively measured how feature importance changes under domain shift. The analysis reveals asymmetric shifts in the influence of key permissions. Fig.~\ref{fig:importance_shifts} presents the mean absolute SHAP values for the top features, comparing intra-domain (blue bars) and cross-domain (orange bars) importance.

In the Pd $\rightarrow$ Nd direction (Fig.~\ref{fig:importance_shifts}a), the importance of \texttt{READ\_PHONE\_STATE} increases substantially compared to its intra-Pd importance. Permissions like \texttt{c2dm.permission.RECEIVE} and \texttt{RECEIVE\_BOOT\_COMPLETED} also show increased influence. Conversely, SMS-related permissions (\texttt{SEND\_SMS}, \texttt{RECEIVE\_SMS}) lose prominence, as they are not among the top indicators in the Nd domain.

In the Nd $\rightarrow$ Pd direction (Fig.~\ref{fig:importance_shifts}b), the predictive power of \texttt{READ\_PHONE\_STATE} decreases when the model encounters the different feature distribution of Pd. Meanwhile, permissions like \texttt{READ\_EXTERNAL\_STORAGE} and \texttt{c2dm.permission.RECEIVE} gain influence. Notably, \texttt{SEND\_SMS} became more important in this cross-direction as it ranks among the top three permissions in the Pd domain.

These direction-dependent changes show that the fitted models' attribution patterns vary across the evaluated populations and provide descriptive evidence of transfer failure.

\subsection{Instance-Level Analysis of Classification Behavior and Failure Modes}

\begin{figure*}[!t]
\centering
\includegraphics[width=\textwidth]{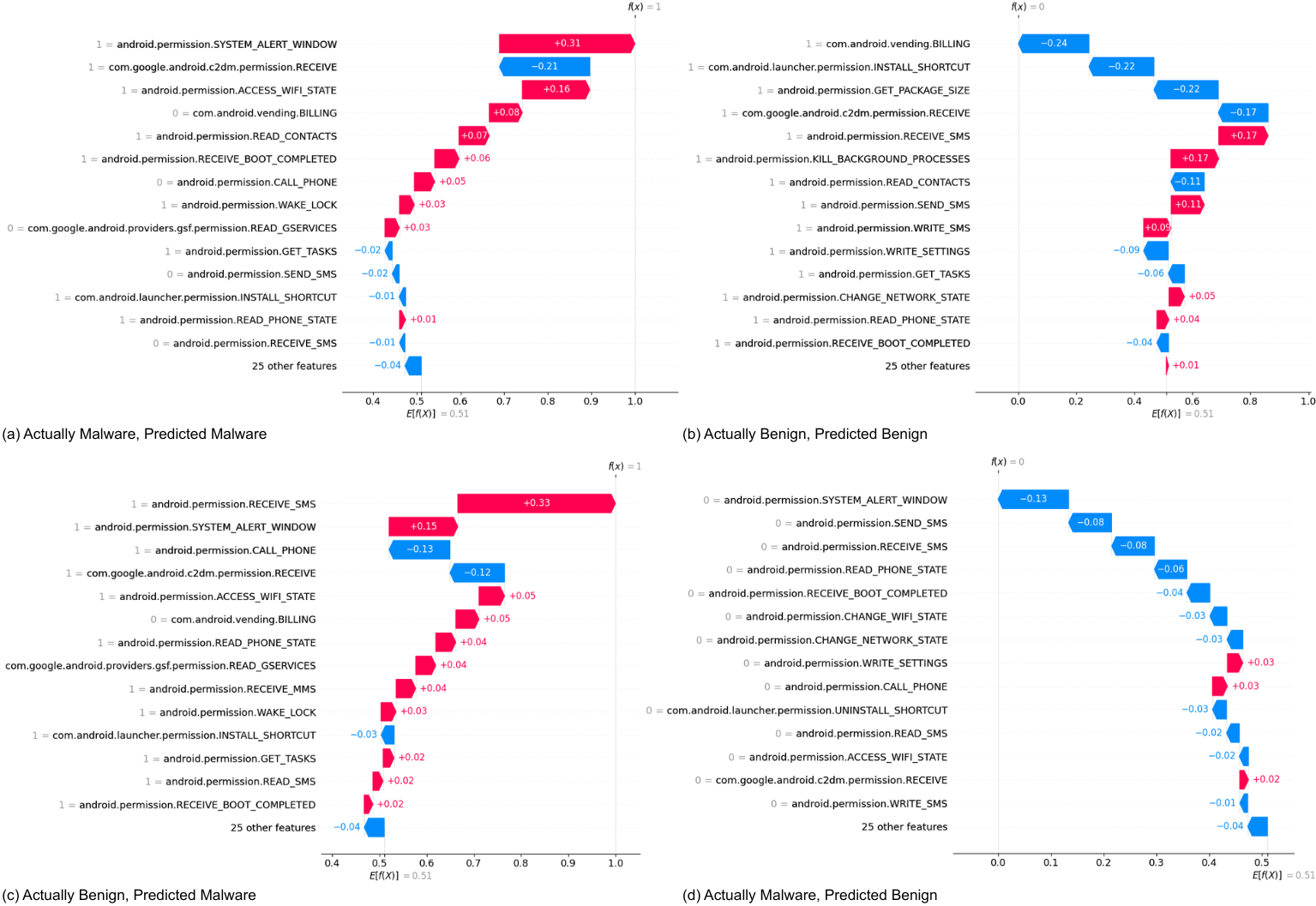}
\caption{SHAP waterfall plots for local interpretation of four instances from the PerMalDroid test set. (a) True positive, (b) true negative, (c) false positive, (d) false negative.}
\label{fig:shap_local}
\end{figure*}

SHAP waterfall diagrams have been applied in four separate cases from the Pd, these involving malware classification (Fig.~\ref{fig:shap_local}a), benign classification (Fig.~\ref{fig:shap_local}b), and two instances of misclassification (Fig.~\ref{fig:shap_local}c and Fig.~\ref{fig:shap_local}d). An examination of these particular examples shows the context-dependent character of permission-based detection. The prediction $f(x)$ is displayed at the top and represents the model's final decision. Permissions with blue bars move the prediction towards 0 (Benign) by reducing the probability, while those with red bars have the opposite effect and drive the prediction towards 1 (Malware). The baseline value, $E[f(X)]$, is the average or expected prediction of the model over the whole training dataset when no information regarding the specific features of the instance is known.

In the correct malware classification (Fig.~\ref{fig:shap_local}a), the prediction is overwhelmingly driven by a few high-risk permissions. The presence of \texttt{SYSTEM\_ALERT\_WINDOW} and \texttt{ACCESS\_WIFI} \texttt{\_STATE} provides the largest positive pushes and clear justification of the model's ``malware'' verdict. The correct benign classification (Fig.~\ref{fig:shap_local}b) shows a different pattern. While the app has some sensitive permissions (\texttt{BILLING}, \texttt{INSTALL\_SHORTCUT}), the model correctly interprets that the combination and context are not malicious because the cumulative evidence is not enough to push the score significantly above the base value.

The misclassifications are highly informative. The false positive (Fig.~\ref{fig:shap_local}c) occurs because the model overvalues the combined presence of \texttt{RECEIVE\_SMS} and \texttt{SYSTEM\_ALERT\_WINDOW} and interprets a potentially legitimate app as malicious. In the false negative case (Fig.~\ref{fig:shap_local}d), where the malware sample successfully evaded detection by either possessing a permission profile that is a complete outlier compared to the training data or, more simply, by not requesting any of the permissions the model relates to malicious behavior.
\subsection{Cross-Dataset Hybrid Training Results}

Table~\ref{tab:hybrid_results} compares hybrid training with source-only and target-only controls. Full hybrid training achieved 85.36--88.01\% mean accuracy on Pd and 96.57--97.01\% on Nd, improving over source-only training by approximately 15--16 and 10--14 percentage points, respectively. However, target-only training achieved 90.58--91.69\% on Pd and 96.64--97.03\% on Nd.

On Pd, full hybrid training underperformed target-only training by 3.42--6.34 percentage points; all paired 95\% bootstrap intervals for these differences lay below zero. Size-matched hybrid training also underperformed target-only training by 2.40--3.68 percentage points, with intervals below zero for all classifiers. This comparison holds total training size constant but replaces approximately half the target-training examples with source examples, so it does not isolate source influence from reduced target-data coverage.

On the Nd, the differences achieved by full hybrid training as compared to target-only training were small (ranging from approximately $-0.18$ to $-0.02$ percentage points). For RF, CB, and HGB, the paired confidence intervals included zero, whereas for XGB and LGBM there were slight decreases. This shows that hybrid training recovers performance relative to source-only transfer but does not offer any advantage over target-only training under the conditions that have been evaluated. The bootstrap intervals are pointwise and based on the fitted models and the evaluation folds.

\begin{table*}[t]
\caption{Hybrid Training Controls: Accuracy (\%, Mean $\pm$ Standard Deviation Across Five Target Folds).}
\label{tab:hybrid_results}
\centering
\small
\setlength{\tabcolsep}{5pt}
\renewcommand{\arraystretch}{1.1}
\begin{tabular}{llcccc}
\toprule
\textbf{Target} & \textbf{Classifier} &
\textbf{Source Only} & \textbf{Target Only} &
\textbf{Full Hybrid} & \textbf{Size-Matched Hybrid} \\
\midrule
Pd & RF   & $72.94 \pm 3.94$ & $91.61 \pm 1.31$ & $88.01 \pm 3.44$ & $89.21 \pm 1.91$ \\
   & XGB  & $70.71 \pm 3.52$ & $91.52 \pm 2.20$ & $86.55 \pm 3.64$ & $88.18 \pm 2.40$ \\
   & LGBM & $69.78 \pm 3.24$ & $91.52 \pm 1.59$ & $85.78 \pm 3.62$ & $87.84 \pm 2.06$ \\
   & CB   & $71.06 \pm 3.67$ & $90.58 \pm 1.90$ & $87.15 \pm 2.45$ & $87.75 \pm 2.15$ \\
   & HGB  & $69.35 \pm 4.04$ & $91.69 \pm 1.57$ & $85.36 \pm 2.95$ & $88.44 \pm 2.31$ \\
\midrule
Nd & RF   & $86.58 \pm 0.54$ & $97.03 \pm 0.27$ & $97.01 \pm 0.30$ & $96.91 \pm 0.32$ \\
   & XGB  & $86.87 \pm 0.56$ & $96.94 \pm 0.37$ & $96.82 \pm 0.36$ & $96.78 \pm 0.34$ \\
   & LGBM & $84.89 \pm 0.56$ & $96.75 \pm 0.35$ & $96.57 \pm 0.43$ & $96.58 \pm 0.44$ \\
   & CB   & $84.46 \pm 0.73$ & $96.99 \pm 0.38$ & $96.95 \pm 0.40$ & $96.95 \pm 0.40$ \\
   & HGB  & $83.06 \pm 0.65$ & $96.64 \pm 0.37$ & $96.61 \pm 0.40$ & $96.55 \pm 0.42$ \\
\bottomrule
\end{tabular}
\end{table*}

\subsection{CORAL-Based Domain Adaptation Results}

The results presented in Table~\ref{tab:coral_results} show how CORAL performs with $\lambda = 10^{-3}$, when compared to similar baselines without CORAL. In the case of Pd $\rightarrow$ Nd, CB's accuracy rose from 84.46\% to 90.21\%, and that of HGB increased from 83.06\% to 87.53\%. RF, XGB, and LGBM all declined; RF, for instance, fell from 86.58\% to 73.56\%, a drop of 13.01 percentage points. This kind of comparison highlights the effect of the transformation when both use the same feature space and the same source training set.

In the Nd $\rightarrow$ Pd direction, XGB, LGBM, CB, and HGB improved by 3.51--5.05 percentage points, while RF changed little (72.94\% to 73.03\%). Although these accuracies remain below the previously reported hybrid results, hybrid training uses labeled examples from both domains and therefore represents a different supervision setting.

The results of the sensitivity analysis presented in Table~\ref{tab:coral_sensitivity} indicate that the outcomes are influenced by both the choice of classifier and the regularization setting. For the RF model, the accuracy stayed below that of its corresponding Pd $\rightarrow$ Nd baseline at all values of $\lambda$, with the highest mean accuracy of 81.84\% being achieved at $\lambda = 0.1$. Therefore, the decline observed in RF was not limited to the case when $\lambda = 10^{-3}$. The CB model showed improvement in this regard at $\lambda = 10^{-4}$ to $10^{-1}$, but suffered a sharp deterioration when $\lambda = 1$. In the opposite direction, the boosting models generally gained from the regularization at $\lambda = 10^{-4}$ to $10^{-1}$, while the RF model exhibited only minor improvements. These results show that CORAL performance depends on both the classifier and the covariance regularization setting.

\begin{table*}[t]
\caption{Matched No-CORAL and CORAL Results at $\lambda = 10^{-3}$ (Mean $\pm$ Standard Deviation). Accuracy is reported in percent.}
\label{tab:coral_results}
\centering
\small
\setlength{\tabcolsep}{5pt}
\renewcommand{\arraystretch}{1.1}
\begin{tabular}{llcccc}
\toprule
\multirow{2}{*}{\textbf{Classifier}} &
\multirow{2}{*}{\textbf{Method}} &
\multicolumn{2}{c}{\textbf{Pd $\rightarrow$ Nd}} &
\multicolumn{2}{c}{\textbf{Nd $\rightarrow$ Pd}} \\
\cmidrule(lr){3-4} \cmidrule(lr){5-6}
& & \textbf{Accuracy} & \textbf{AUC} & \textbf{Accuracy} & \textbf{AUC} \\
\midrule
RF & No CORAL & $86.58 \pm 0.54$ & $0.922 \pm 0.005$ & $72.94 \pm 3.94$ & $0.837 \pm 0.040$ \\
   & CORAL & $73.56 \pm 7.12$ & $0.912 \pm 0.014$ & $73.03 \pm 3.55$ & $0.837 \pm 0.022$ \\
\midrule
XGB & No CORAL & $86.87 \pm 0.56$ & $0.939 \pm 0.004$ & $70.71 \pm 3.52$ & $0.773 \pm 0.049$ \\
    & CORAL & $81.55 \pm 10.02$ & $0.935 \pm 0.009$ & $74.22 \pm 3.24$ & $0.826 \pm 0.037$ \\
\midrule
LGBM & No CORAL & $84.89 \pm 0.56$ & $0.927 \pm 0.004$ & $69.78 \pm 3.24$ & $0.764 \pm 0.049$ \\
     & CORAL & $82.08 \pm 4.73$ & $0.936 \pm 0.008$ & $74.05 \pm 2.49$ & $0.815 \pm 0.039$ \\
\midrule
CB & No CORAL & $84.46 \pm 0.73$ & $0.941 \pm 0.004$ & $71.06 \pm 3.67$ & $0.780 \pm 0.045$ \\
   & CORAL & $90.21 \pm 1.00$ & $0.949 \pm 0.004$ & $76.11 \pm 3.15$ & $0.842 \pm 0.025$ \\
\midrule
HGB & No CORAL & $83.06 \pm 0.65$ & $0.914 \pm 0.005$ & $69.35 \pm 4.04$ & $0.763 \pm 0.048$ \\
    & CORAL & $87.53 \pm 0.80$ & $0.941 \pm 0.006$ & $74.05 \pm 2.16$ & $0.821 \pm 0.032$ \\
\bottomrule
\end{tabular}
\end{table*}

\begin{table*}[t]
\caption{CORAL Regularization Sensitivity: Accuracy (\%, Mean $\pm$ Standard Deviation). No-CORAL baselines are reported in Table~\ref{tab:coral_results}.}
\label{tab:coral_sensitivity}
\centering
\small
\setlength{\tabcolsep}{5pt}
\renewcommand{\arraystretch}{1.1}
\begin{tabular}{llccccc}
\toprule
\textbf{Direction} & \textbf{Classifier} &
$\boldsymbol{\lambda=10^{-4}}$ &
$\boldsymbol{\lambda=10^{-3}}$ &
$\boldsymbol{\lambda=10^{-2}}$ &
$\boldsymbol{\lambda=10^{-1}}$ &
$\boldsymbol{\lambda=1}$ \\
\midrule
Pd $\rightarrow$ Nd & RF & $73.31 \pm 5.46$ & $73.56 \pm 7.12$ & $78.25 \pm 8.09$ & $81.84 \pm 7.38$ & $74.97 \pm 10.95$ \\
& XGB & $77.89 \pm 9.98$ & $81.55 \pm 10.02$ & $87.69 \pm 2.97$ & $72.13 \pm 10.65$ & $75.73 \pm 8.24$ \\
& LGBM & $84.31 \pm 6.12$ & $82.08 \pm 4.73$ & $81.84 \pm 7.71$ & $80.40 \pm 11.36$ & $68.26 \pm 4.42$ \\
& CB & $90.22 \pm 0.69$ & $90.21 \pm 1.00$ & $89.15 \pm 3.50$ & $88.37 \pm 2.26$ & $58.36 \pm 8.93$ \\
& HGB & $82.73 \pm 7.00$ & $87.53 \pm 0.80$ & $83.55 \pm 8.01$ & $81.72 \pm 6.14$ & $74.37 \pm 2.65$ \\
\midrule
Nd $\rightarrow$ Pd & RF & $72.17 \pm 4.21$ & $73.03 \pm 3.55$ & $72.17 \pm 3.10$ & $72.25 \pm 4.09$ & $57.70 \pm 5.22$ \\
& XGB & $73.20 \pm 2.46$ & $74.22 \pm 3.24$ & $72.43 \pm 3.40$ & $72.08 \pm 3.94$ & $65.66 \pm 6.36$ \\
& LGBM & $73.63 \pm 2.40$ & $74.05 \pm 2.49$ & $73.88 \pm 3.01$ & $74.05 \pm 2.77$ & $69.18 \pm 3.48$ \\
& CB & $75.51 \pm 3.02$ & $76.11 \pm 3.15$ & $75.94 \pm 3.50$ & $76.37 \pm 3.94$ & $72.94 \pm 5.74$ \\
& HGB & $73.80 \pm 2.10$ & $74.05 \pm 2.16$ & $73.28 \pm 2.19$ & $74.48 \pm 2.63$ & $69.86 \pm 2.29$ \\
\bottomrule
\end{tabular}
\end{table*}

\section{Discussion}

Our experiments reveal several insights into domain shift in permission-based Android malware detection. Models trained on outer training partitions of the 1,168-sample Pd dataset achieved 85--87\% mean accuracy on the unseen Nd dataset containing 29,332 apps. Feature selection was performed independently of the final Pd test folds, and selected feature counts ranged from 23 to 559. Thus, transfer performance persisted across source partitions, although the size of the selected subset was not stable.

The difference in transfer directions is linked to a mismatch in feature coverage, differences in model reliance on permissions, and cross-domain class overlap. Even though permissions that are not included in Nd's vocabulary of 86 features are unavailable to models trained on Nd, their role in causing transfer failure cannot be determined solely on the basis of overlap counts. Analysis using SHAP shows that there are differences in the way permissions contribute across domains. More directly, the high-dimensional nearest-neighbor analysis showed that 59.00\% of the Pd malware samples that were missed were closer to the benign training samples of Nd, as compared with 5.87\% of the malware samples that were correctly detected. This provides support for the association between relative proximity to benign permission profiles and missed detections beyond the exploratory PCA projection.

The ablation study found that feature selection increased the mean Pd $\rightarrow$ Nd accuracy by 4.85 percentage points for RF, 2.11 for XGB, 0.53 for LGBM, and 2.62 for HGB, whereas CB decreased very slightly from 86.09\% to 85.97\%. However, the full-feature models still had higher mean Pd intra-domain accuracy for all the classifiers. These results show that there is a model-dependent trade-off between fit to the source domain and transfer performance, rather than indicating a general need for feature reduction. The findings do not prove that the regularization mechanism of a specific classifier accounts for its reaction to feature selection.

The explainability analysis shows how the model's dependence on permissions varies across the evaluated groups. The changes in the mean absolute SHAP values indicate differences in feature distributions and in how the model assigns attribution to them; such changes should not be taken as direct measures of changes in the process that generates the labels. Practically, the presence of shared permissions alone does not ensure that a detector will use them reliably in a new environment.

Training using a supervised hybrid approach on the 85 common permissions led to a higher level of accuracy than transfer based solely on the source, reaching 85--88\% on Pd and about 97\% on Nd. Yet training only on the target achieved about 91\% on Pd and 97\% on Nd. Full hybrid training performed worse than target-only training on Pd by 3.42 to 6.34 percentage points, with all of the paired 95\% bootstrap intervals being below zero. Size-matched hybrid training also underperformed target-only training, although it replaced approximately half the target examples with source examples. On Nd, differences were small. These controls provide no demonstrated advantage of pooling over target-only training under the evaluated settings and show that access to labeled target data is an important component of the recovery from source-only transfer.

The performance of CORAL varied depending on both the type of classifier and the regularization parameter. When $\lambda$ was set at $10^{-3}$, CB and HGB achieved gains of 5.74 and 4.48 percentage points, respectively, in the transfer from Pd to Nd, while RF lost 13.01 points compared to its corresponding baseline. RF stayed below its baseline at all the values of $\lambda$ tested in this case. For the transfer from Nd to Pd, the boosted classifiers showed improvements of 3.51 to 5.05 points, while RF's paired accuracy-difference interval included zero. These findings show that the results are sensitive to alignment and regularization, but they do not identify the mechanism responsible for the degradation nor determine whether covariance mismatch is a main cause of the domain shift. Moreover, CORAL and hybrid training differ in their access to the target labels and therefore should not be regarded as equivalent supervision scenarios.

There are several limitations that need to be recognized. The study evaluates only two datasets, and its results cannot be extended to other stores, different regions, or other time periods. Permissions used in the study might also change as the Android platform develops. The bootstrap intervals are pointwise, have not been adjusted for multiple comparisons, and are based on the fitted models and the way the data is divided into folds. Therefore, they do not take into account the variability due to retraining or the uncertainty that arises from different data sources. Lastly, permission-only representations are unable to distinguish between malware and benign applications, as shown by the exact matches in the nearest-neighbor analysis. Further research into complementary static and dynamic features should be carried out.

\section{Conclusion}

The main contributions of this paper are threefold. We first analyzed domain shift in permission-based malware detection using two distinct datasets. With feature selection performed independently of the final Pd test folds, models achieved 89--90\% mean intra-domain accuracy on Pd and 85--87\% in Pd $\rightarrow$ Nd transfer. In contrast, Nd-trained models achieved approximately 97\% intra-domain accuracy but only 69--73\% on Pd, a reduction of approximately 24--28 percentage points.

Second, explainable AI together with a high-dimensional nearest-neighbor analysis was used to examine transfer failure. SHAP showed that the contributions of permission features varied across different domains, and the nearest-neighbor analysis found that the Pd malware samples that had been missed were more often found to be nearer to Nd benign training samples than correctly detected malware samples were. These results indicate an association between cross-domain class overlap and failed detections. In the case of the matched ablation, the mean transfer accuracy increased for four out of five of the classifiers, even though the number of selected features differed between the source partitions and the classifiers.

Third, supervised hybrid training using common permissions achieved 85--88\% accuracy on Pd and about 97\% on Nd. The matched control experiments showed an improvement compared with source-only transfer but offered no advantage over target-only training, thus highlighting the role of the labeled target examples. The CORAL results depended on both the choice of classifier and the regularization setting, with the CB classifier achieving 90.21\% accuracy at $\lambda = 10^{-3}$ in the Pd $\rightarrow$ Nd transfer scenario. When the direction was reversed, boosting the classifiers resulted in gains of 3.51 to 5.05 percentage points at this setting, whereas RF showed no clear improvement in accuracy. However, these results on their own do not demonstrate the relative contribution of covariance mismatch to transfer failure.

Further research should look into more advanced adaptation methods, temporal drift, and additional independent datasets. The findings of this paper are specific to the evaluated dataset pair, and the bootstrap intervals are conditional on the fitted models and folds. Classifier performance could be further improved by combining permission features with complementary static and dynamic signals.

\section*{Declaration of Generative AI and AI-Assisted
Technologies in the Manuscript Preparation Process}

During the preparation of this manuscript, the authors used ChatGPT (OpenAI) to assist with language refinement, improving clarity, LaTeX codes and manuscript formatting. The authors reviewed and edited the output as needed and take full responsibility for the content of the published article.

\appendix
\section{Confidence Intervals for Control Experiments}
\label{app:confidence_intervals}

Tables~\ref{tab:coral_ci}, \ref{tab:paired_ci}, and \ref{tab:hybrid_ci} report pooled estimates and pointwise 95\% bootstrap confidence intervals. Intervals are conditional
on the fitted models and folds and are not adjusted for multiple comparisons.

\clearpage

\begin{table*}[!tp]
\caption{CORAL and Matched Baseline: Accuracy and Malware Recall (\%) with 95\% Confidence Intervals. CORAL uses $\lambda=10^{-3}$.}
\label{tab:coral_ci}
\centering
\small
\setlength{\tabcolsep}{5pt}
\renewcommand{\arraystretch}{1.1}
\begin{tabular}{llcccc}
\toprule
\multirow{2}{*}{\textbf{Direction}} &
\multirow{2}{*}{\textbf{Classifier}} &
\multicolumn{2}{c}{\textbf{No CORAL}} &
\multicolumn{2}{c}{\textbf{CORAL}} \\
\cmidrule(lr){3-4} \cmidrule(lr){5-6}
& & \textbf{Accuracy} & \textbf{Malware Recall}
& \textbf{Accuracy} & \textbf{Malware Recall} \\
\midrule
Pd $\rightarrow$ Nd & RF
& 86.58 [86.20, 86.96] & 82.95 [82.34, 83.56]
& 73.56 [73.12, 74.00] & 54.44 [53.65, 55.21] \\
& XGB
& 86.87 [86.48, 87.25] & 83.06 [82.44, 83.66]
& 81.55 [81.16, 81.94] & 70.91 [70.27, 71.54] \\
& LGBM
& 84.89 [84.50, 85.30] & 80.10 [79.46, 80.74]
& 82.08 [81.66, 82.49] & 71.36 [70.64, 72.06] \\
& CB
& 84.46 [84.06, 84.87] & 76.59 [75.91, 77.28]
& 90.21 [89.87, 90.54] & 89.74 [89.25, 90.22] \\
& HGB
& 83.06 [82.63, 83.48] & 78.01 [77.33, 78.65]
& 87.53 [87.16, 87.90] & 83.35 [82.75, 83.95] \\
\midrule
Nd $\rightarrow$ Pd & RF
& 72.95 [70.55, 75.34] & 59.47 [55.65, 63.12]
& 73.03 [70.72, 75.43] & 56.98 [52.99, 60.96] \\
& XGB
& 70.72 [68.24, 73.12] & 55.32 [51.49, 59.30]
& 74.23 [71.83, 76.54] & 61.13 [57.31, 64.95] \\
& LGBM
& 69.78 [67.30, 72.26] & 54.32 [50.50, 58.31]
& 74.06 [71.58, 76.46] & 62.79 [58.97, 66.61] \\
& CB
& 71.06 [68.66, 73.46] & 55.81 [51.99, 59.80]
& 76.11 [73.72, 78.51] & 69.77 [66.11, 73.42] \\
& HGB
& 69.35 [66.95, 71.83] & 52.82 [48.84, 56.81]
& 74.06 [71.58, 76.46] & 62.96 [59.14, 66.78] \\
\bottomrule
\end{tabular}
\end{table*}

\begin{table*}[!tp]
\caption{Paired Accuracy Differences in Percentage Points with 95\% Confidence Intervals. C: CORAL ($\lambda=10^{-3}$) minus no-CORAL baseline; H--S: full hybrid minus source-only; H--T: full hybrid minus target-only; M--T: size-matched hybrid minus target-only. Positive values favor the first method.}
\label{tab:paired_ci}
\centering
\small
\setlength{\tabcolsep}{4pt}
\renewcommand{\arraystretch}{1.1}
\begin{tabular}{llcccc}
\toprule
\textbf{Direction} & \textbf{Classifier} &
\textbf{C} & \textbf{H--S} & \textbf{H--T} & \textbf{M--T} \\
\midrule
Pd $\rightarrow$ Nd & RF
& $-13.01$ [$-13.45$, $-12.56$]
& $10.43$ [$10.06$, $10.81$]
& $-0.02$ [$-0.09$, $0.04$]
& $-0.12$ [$-0.20$, $-0.04$] \\
& XGB
& $-5.32$ [$-5.75$, $-4.90$]
& $9.95$ [$9.57$, $10.33$]
& $-0.12$ [$-0.22$, $-0.03$]
& $-0.16$ [$-0.25$, $-0.07$] \\
& LGBM
& $-2.81$ [$-3.25$, $-2.37$]
& $11.68$ [$11.29$, $12.09$]
& $-0.18$ [$-0.28$, $-0.10$]
& $-0.17$ [$-0.26$, $-0.08$] \\
& CB
& $5.74$ [$5.38$, $6.11$]
& $12.49$ [$12.09$, $12.90$]
& $-0.04$ [$-0.12$, $0.03$]
& $-0.04$ [$-0.12$, $0.03$] \\
& HGB
& $4.48$ [$4.07$, $4.88$]
& $13.56$ [$13.13$, $13.99$]
& $-0.02$ [$-0.12$, $0.07$]
& $-0.09$ [$-0.19$, $0.01$] \\
\midrule
Nd $\rightarrow$ Pd & RF
& $0.09$ [$-2.14$, $2.40$]
& $15.07$ [$13.10$, $17.12$]
& $-3.60$ [$-5.39$, $-1.88$]
& $-2.40$ [$-3.85$, $-0.94$] \\
& XGB
& $3.51$ [$1.11$, $5.82$]
& $15.84$ [$13.87$, $17.89$]
& $-4.97$ [$-6.94$, $-3.00$]
& $-3.34$ [$-5.05$, $-1.63$] \\
& LGBM
& $4.28$ [$2.14$, $6.51$]
& $16.01$ [$14.04$, $18.07$]
& $-5.74$ [$-7.62$, $-3.85$]
& $-3.68$ [$-5.22$, $-2.14$] \\
& CB
& $5.05$ [$2.74$, $7.28$]
& $16.10$ [$14.04$, $18.15$]
& $-3.42$ [$-5.14$, $-1.71$]
& $-2.83$ [$-4.28$, $-1.37$] \\
& HGB
& $4.71$ [$2.57$, $6.76$]
& $16.01$ [$14.04$, $18.07$]
& $-6.34$ [$-8.31$, $-4.37$]
& $-3.25$ [$-4.71$, $-1.80$] \\
\bottomrule
\end{tabular}
\end{table*}
\clearpage
\begin{table*}[!tp]
\caption{Hybrid Controls: Accuracy and Malware Recall (\%) with 95\% Confidence Intervals.}
\label{tab:hybrid_ci}
\centering
\small
\setlength{\tabcolsep}{5pt}
\renewcommand{\arraystretch}{1.05}
\begin{tabular}{lllll}
\toprule
\textbf{Target} & \textbf{Classifier} & \textbf{Condition}
& \textbf{Accuracy} & \textbf{Malware Recall} \\
\midrule
Nd & RF & Source only & 86.58 [86.19, 86.96] & 82.95 [82.34, 83.56] \\
& & Target only & 97.03 [96.84, 97.23] & 96.54 [96.24, 96.84] \\
& & Full hybrid & 97.01 [96.81, 97.20] & 96.53 [96.23, 96.82] \\
& & Size-matched hybrid & 96.91 [96.72, 97.11] & 96.42 [96.12, 96.71] \\
\cmidrule(lr){2-5}
& XGB & Source only & 86.87 [86.50, 87.26] & 83.06 [82.46, 83.68] \\
& & Target only & 96.94 [96.74, 97.14] & 96.60 [96.30, 96.88] \\
& & Full hybrid & 96.82 [96.62, 97.02] & 96.59 [96.30, 96.88] \\
& & Size-matched hybrid & 96.78 [96.58, 96.98] & 96.57 [96.27, 96.86] \\
\cmidrule(lr){2-5}
& LGBM & Source only & 84.89 [84.49, 85.30] & 80.10 [79.46, 80.74] \\
& & Target only & 96.75 [96.55, 96.95] & 96.69 [96.40, 96.97] \\
& & Full hybrid & 96.57 [96.36, 96.78] & 96.70 [96.41, 96.99] \\
& & Size-matched hybrid & 96.58 [96.37, 96.79] & 96.74 [96.44, 97.03] \\
\cmidrule(lr){2-5}
& CB & Source only & 84.46 [84.06, 84.85] & 76.59 [75.90, 77.26] \\
& & Target only & 96.99 [96.79, 97.18] & 96.76 [96.46, 97.04] \\
& & Full hybrid & 96.95 [96.75, 97.14] & 96.77 [96.48, 97.05] \\
& & Size-matched hybrid & 96.95 [96.75, 97.14] & 96.76 [96.47, 97.04] \\
\cmidrule(lr){2-5}
& HGB & Source only & 83.06 [82.61, 83.48] & 78.01 [77.34, 78.67] \\
& & Target only & 96.64 [96.43, 96.84] & 96.50 [96.20, 96.79] \\
& & Full hybrid & 96.62 [96.41, 96.82] & 96.65 [96.35, 96.93] \\
& & Size-matched hybrid & 96.55 [96.35, 96.76] & 96.74 [96.45, 97.02] \\
\midrule
Pd & RF & Source only & 72.95 [70.55, 75.34] & 59.47 [55.64, 63.29] \\
& & Target only & 91.61 [89.98, 93.15] & 92.36 [90.20, 94.52] \\
& & Full hybrid & 88.01 [86.22, 89.81] & 85.38 [82.56, 88.04] \\
& & Size-matched hybrid & 89.21 [87.41, 91.01] & 90.70 [88.37, 93.02] \\
\cmidrule(lr){2-5}
& XGB & Source only & 70.72 [68.32, 73.12] & 55.32 [51.49, 59.30] \\
& & Target only & 91.52 [89.90, 93.07] & 91.86 [89.54, 93.85] \\
& & Full hybrid & 86.56 [84.59, 88.44] & 85.05 [82.23, 87.71] \\
& & Size-matched hybrid & 88.19 [86.30, 89.98] & 88.70 [86.21, 91.20] \\
\cmidrule(lr){2-5}
& LGBM & Source only & 69.78 [67.38, 72.18] & 54.32 [50.33, 58.14] \\
& & Target only & 91.52 [89.90, 93.07] & 91.53 [89.20, 93.69] \\
& & Full hybrid & 85.79 [83.82, 87.76] & 84.39 [81.56, 87.21] \\
& & Size-matched hybrid & 87.84 [85.96, 89.73] & 89.20 [86.71, 91.69] \\
\cmidrule(lr){2-5}
& CB & Source only & 71.06 [68.58, 73.46] & 55.81 [51.83, 59.80] \\
& & Target only & 90.58 [88.87, 92.21] & 90.03 [87.71, 92.36] \\
& & Full hybrid & 87.16 [85.19, 89.04] & 85.71 [82.89, 88.37] \\
& & Size-matched hybrid & 87.76 [85.87, 89.56] & 89.20 [86.71, 91.53] \\
\cmidrule(lr){2-5}
& HGB & Source only & 69.35 [66.95, 71.75] & 52.82 [49.00, 56.65] \\
& & Target only & 91.70 [90.07, 93.24] & 91.86 [89.70, 94.02] \\
& & Full hybrid & 85.36 [83.31, 87.33] & 83.72 [80.73, 86.55] \\
& & Size-matched hybrid & 88.44 [86.56, 90.24] & 89.37 [86.88, 91.69] \\
\bottomrule
\end{tabular}
\end{table*}

\end{document}